\documentclass[runningheads]{llncs}
\usepackage{graphicx}
\usepackage{amsmath,amssymb} 
\usepackage{color}
\usepackage[width=122mm,left=12mm,paperwidth=146mm,height=193mm,top=12mm,paperheight=217mm]{geometry}

\usepackage{times}

\usepackage{pstricks}
\usepackage{xspace}            
\usepackage{multirow}
\usepackage{subfig}
\usepackage{multirow}
\usepackage{array}
\usepackage{url}
\usepackage{pdfpages}
\usepackage{euscript}

\usepackage{mathtools}
\usepackage{enumitem}
\usepackage{epstopdf}
\usepackage{float}
\usepackage[linesnumbered,ruled,vlined]{algorithm2e}
\usepackage{wrapfig}
\usepackage{bbold}

\usepackage{url}

\newcommand{\mat}[1]{\mathtt #1}

\newcommand{\ind}[1]{\mathbb 1_{#1}}

\newcommand{\vct}[1]{\mathbf #1}

\newcommand{\argmin}{\operatornamewithlimits{\arg\,\min}}

\newcommand{\x}{{\bf x}}

\def\QEDopen{{\setlength{\fboxsep}{0pt}\setlength{\fboxrule}{0.2pt}\fbox{\rule[0pt]{0pt}{1.3ex}\rule[0pt]{1.3ex}{0pt}}}}
\def\QED{\QEDopen} 
\def\proof{\noindent\hspace{1.1em}{\itshape Proof. }}
\def\endproof{\hspace*{\fill}~\QED~~~\par\endtrivlist\unskip\medskip}

\begin{document}

\pagestyle{headings}
\mainmatter

\title{Interactive Image Segmentation Using\\Constrained Dominant Sets} 

\titlerunning{Interactive image segmentation using constrained dominant sets}

\authorrunning{E. Zemene and M. Pelillo}

\author{Eyasu Zemene and Marcello Pelillo}

\institute{Ca' Foscari University of Venice, Italy\\
\email{ \{eyasu.zemene,pelillo\}@unive.it}}
\maketitle

\begin{abstract}
We propose a new approach to interactive image segmentation based on some properties of a family of
quadratic optimization problems related to dominant sets, a well-known graph-theoretic notion of a cluster
which generalizes the concept of a maximal clique to edge-weighted graphs. In particular, we show that by properly controlling a regularization parameter which determines the structure and the scale of the underlying problem, we are in a position to extract groups of dominant-set clusters which are constrained to contain user-selected elements. The resulting algorithm can deal naturally with any type of input modality, including scribbles, sloppy contours, and bounding boxes, and is able to robustly handle noisy annotations on the part of the user. Experiments on standard benchmark datasets show the effectiveness of our approach as compared to state-of-the-art algorithms on a variety of natural images under several input conditions.
\end{abstract}

\section{Introduction}

User-assisted image segmentation has recently attracted considerable attention within the computer vison community, especially because of its potential applications in a variety of different problems such as image and video editing, medical image analysis, etc. \cite{GrabCutRotherKB04,iccvLempitsky09,MilCutCVPR14,BaiSapIJCV2009,LiSunTanShuACM2004,ProSapIP2007,BoyJolICCV2001,MorBarIP1998}.
Given an input image and some information provided by a user, usually in the form of a scribble or of a bounding box,
the goal is to provide as output a foreground object in such a way as to best reflect the user's intent.
By exploiting high-level, semantic knowledge on the part of the user, which is typically difficult to formalize, we are therefore able to effectively solve segmentation problems which would be otherwise too complex to be tackled using fully automatic segmentation algorithms.

Existing algorithms fall into two broad categories, depending on whether the user annotation is given in terms of a scribble or of a bounding box, and supporters of the two approaches have both good reasons to prefer one
modality against the other.
For example, Wu et al. \cite{MilCutCVPR14} claim that bounding boxes are the most natural and economical form
in terms of the amount of user interaction, and develop a multiple instance learning algorithm that extracts an arbitrary object located inside a tight bounding box at unknown location. 
Yu et al. \cite{LOOSECUTcorr15} also support the bounding-box approach, though their algorithm is different from others in that it does not need bounding boxes tightly enclosing the object of interest, whose production of course increases the annotation burden. They provide an algorithm, based on a Markov Random Field (MRF) energy function, that can handle input bounding box that only loosely covers the foreground object. 
Xian et al. \cite{XiaZhaCheXuDinCoRR2015} propose a method which avoids the limitations of existing bounding box methods - region of interest (ROI) based methods, though they need much less user interaction, their performance is sensitive to initial ROI. 
On the other hand, several researchers, arguing that boundary-based interactive segmentation such as intelligent scissors \cite{MorBarIP1998} requires the user to trace the whole boundary of the object, which is usually a time-consuming and tedious process, support scribble-based segmentation. Bai et al. \cite{BaiWuCVPR2014}, for example, propose a model based on ratio energy function which can be optimized using an iterated graph cut algorithm, which tolerates errors in the user input.

In general, the input modality in an interactive segmentation algorithm affects both its accuracy and its ease of use. Existing methods work typically on a single modality and they focus on how to use that input most effectively. However, as noted recently by Jain and Grauman \cite{JainGraICCV2013}, sticking to one annotation form leads to a suboptimal tradeoff between human and machine effort, and they tried to estimate how much user input is required to sufficiently segment a novel input.

\begin{figure}[t]
\centering
\includegraphics[height=4.0cm,width=8cm]{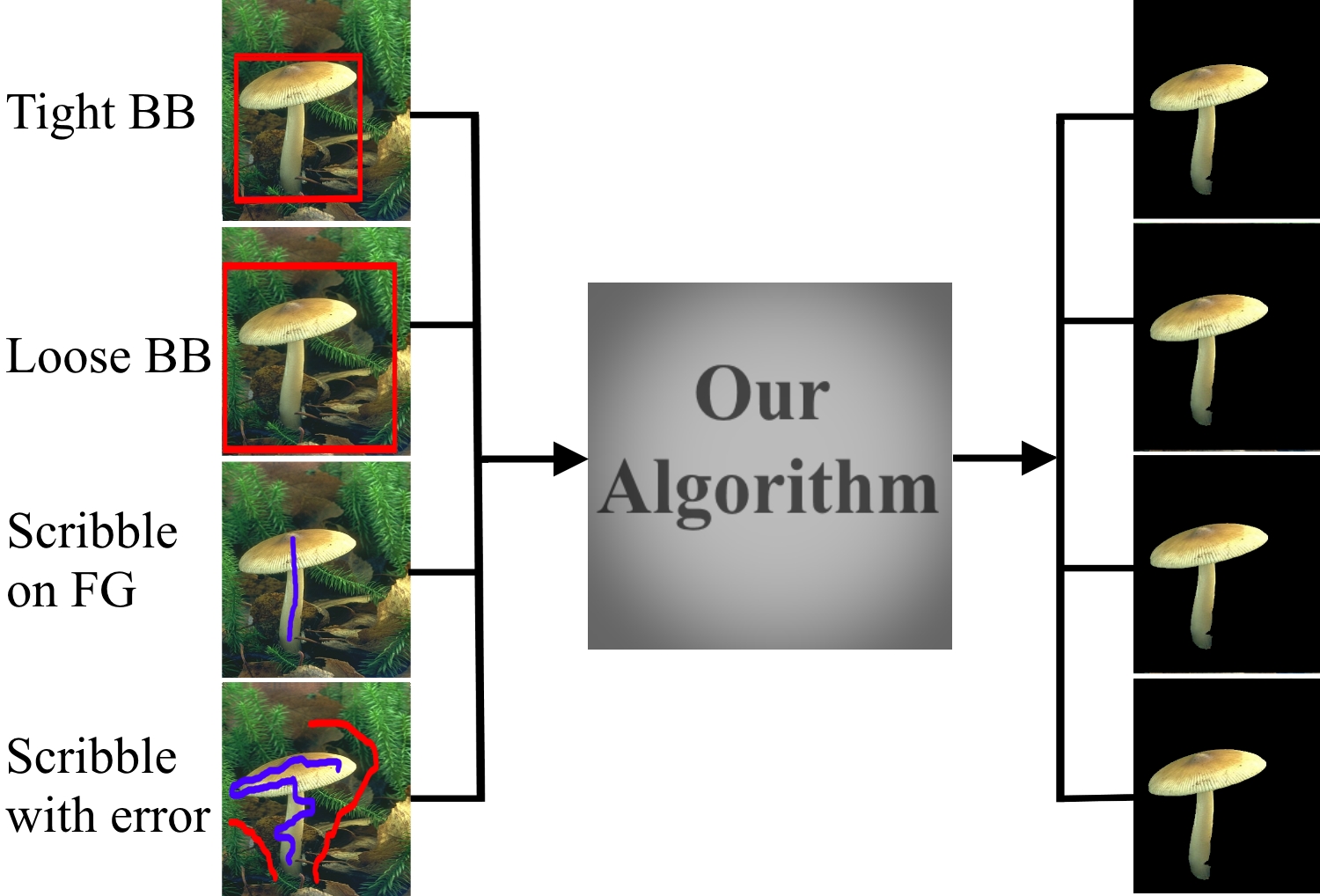}\\
\caption{\small{ \bf Left:} An input image with different user annotations.
Tight bounding box (Tight BB), loose bounding box (Loose BB),
a scribble made (only) on the foreground object (Scribbles on FG), scribbles with errors.
{\bf Right:} Results of the proposed algorithm.}
\label{fig:InputModalities}
\end{figure}

In this paper, we propose a novel approach to interactive image segmentation which can deal naturally with any type of input modality and is able to robustly handle noisy annotations on the part of the user. Figure \ref{fig:InputModalities} shows an example of how our system works in the presence of different input annotations. 
Our approach is based on some properties of a parameterized family of quadratic optimization problems related to dominant-set clusters, a well-known generalization of the notion of maximal cliques to edge-weighted graph which have proven to be extremely effective in a variety of computer vision problems, including (automatic) image and video segmentation \cite{PavPelCVPR2003,PavPel07}.
In particular, we show that by properly controlling a regularization parameter which determines the structure
and the scale of the underlying problem, we are in a position to extract groups of dominant-set clusters which
are constrained to contain user-selected elements. We provide bounds that allow us to control this process,
which are based on the spectral properties of certain submatrices of the original affinity matrix.

The resulting algorithm has a number of interesting features which distinguishes it from existing approaches.
Specifically: 1) it is able to deal in a flexible manner with {\em both} scribble-based and
boundary-based input modalities (such as sloppy contours and bounding boxes); 2) in the case of noiseless scribble inputs, it asks the user to provide {\em only} foreground pixels; 3) it turns out to be {\em robust} in the presence of input noise, allowing the user to draw, e.g., imperfect scribbles (including background pixels) or loose bounding boxes.
Experimental results on standard benchmark datasets show the effectiveness of our approach as compared to state-of-the-art algorithms on a wide variety of natural images under several input conditions.

\section{Dominant sets and quadratic optimization}
\label{sec:DS}
In this section we review the basic definitions and properties of dominant sets, as introduced in \cite{PavPelCVPR2003,PavPel07}.
In the dominant set framework, the data to be clustered are represented as an undirected edge-weighted graph with no self-loops $G = (V, E,w)$, where $V = \{1, . . . , n\}$ is the vertex set, $E \subseteq V \times V$ is the edge set, and $w : E \rightarrow R_+^*$ is the (positive) weight function. Vertices in $G$ correspond to data points, edges represent neighborhood relationships, and edge-weights reflect similarity between pairs of linked vertices. As customary, we represent the graph $G$ with the corresponding weighted adjacency (or similarity) matrix, which is the $n \times n$ nonnegative, symmetric matrix $A = (a_{ij})$ defined as $a_{ij} = w(i, j)$, if $(i, j) \in E$, and $a_{ij} = 0$ otherwise. Since in $G$ there are no self-loops, note that all entries on the main diagonal of $A$ are zero.

For a non-empty subset $S \subseteq V$, $i \in S$, and $j \notin S$, define
\begin{equation}
\label{eq1}
\phi_S(i,j)=a_{ij}-\frac{1}{|S|} \sum_{k \in S} a_{ik}
\end{equation}

Next, to each vertex $i \in S$ we assign a weight defined (recursively) as follows:
\begin{equation}
w_S(i)=
\begin{cases}
1,&\text{if\quad $|S|=1$},\\
\sum_{j \in S \setminus \{i\}} \phi_{S \setminus \{i\}}(j,i)w_{S \setminus \{i\}}(j),&\text{otherwise}.
\end{cases}
\end{equation}
As explained in \cite{PavPelCVPR2003,PavPel07}, a positive $w_S(i)$ indicates that adding $i$ into its neighbors in $S$ will increase the internal coherence of the set, whereas in the presence of a negative value we expect the overall coherence to be decreased. Finally, the total weight of $S$ can be simply defined as
\begin{equation}
W(S)=\sum_{i \in S}w_S(i)~.
\end{equation}

A non-empty subset of vertices $S \subseteq V$ such that $W(T) > 0$ for any non-empty $T \subseteq S$, is said to be a {\em dominant set} if:
\begin{enumerate}

\item  $w_S(i)>0$, for all $i \in S$,
\item  $w_{S \cup \{i\}}(i)<0$, for all $i \notin S$.
\end{enumerate}
It is evident from the definition that a dominant set satisfies the two basic properties of a cluster: internal coherence and external incoherence. Condition 1 indicates that a dominant set is internally coherent, while condition 2 implies that this coherence will be destroyed by the addition of any vertex from outside. In other words, a dominant set is a maximally coherent data set.

Now, consider the following linearly-constrained quadratic optimization problem:
\begin{equation}
\label{eq2}
\begin{array}{ll}
   \text{maximize }  &  f(\x) = \x' A \x \\
   \text{subject to} &  \mathbf{x} \in \Delta
\end{array}
\end{equation}
where a prime denotes transposition and  
$$
 \Delta=\left\{ \x \in R^n~:~ \sum_{i=1}^n x_i = 1, \text{ and } x_i \geq 0 \text{ for all } i=1 \ldots n \right\}
$$ 
is the standard simplex of $R^n$.
In \cite{PavPelCVPR2003,PavPel07} a connection is established between dominant sets and the local solutions of \eqref{eq2}. In particular, it is shown that if $S$ is a dominant set then its ``weighted characteristics vector,'' which is the vector of $\Delta$ defined as,
\begin{displaymath}
 x_i=
\begin{cases} \frac{w_S(i)}{W(s)},&\text{if\quad $i \in S$},\\ 0,&\text{otherwise}
\end{cases}
\end{displaymath}
is a strict local solution of \eqref{eq2}. Conversely, under mild conditions, it turns out that if $\x$ is a (strict) local solution of program \eqref{eq2} then its ``support''
$$
\sigma(\x) = \{i \in V~:~x_i > 0\}
$$
is a dominant set.
By virtue of this result, we can find a dominant set by first localizing a solution of program \eqref{eq2} with an appropriate continuous optimization technique, and then picking up the support set of the solution found. A generalization of these ideas to hypergraphs has recently been developed in \cite{RotPelPAMI2013}.

A simple and effective optimization algorithm to extract a dominant set from a graph is given by the so-called {\em replicator dynamics}, developed and studied in evolutionary game theory, which are defined as follows:
\begin{equation}
x_i^{(t+1)} = x_i^{(t)} \frac{(A\x^{(t)})_i}{(\x^{(t)})'A(\x^{(t)})}
\label{eqn:Replicator}
\end{equation}
for $i=1,\ldots,n$.

\section{Constrained dominant sets}

Let $G=(V,E,w)$ be an edge-weighted graph with $n$ vertices and let $A$ denote as usual its (weighted) adjacency matrix. Given a subset of vertices $S \subseteq V$ and a parameter $ \alpha > 0$, define
the following parameterized family of quadratic programs:
\begin{equation}
\label{eqn:parQP}
\begin{array}{ll}
   \text{maximize }  &  f_S^\alpha(\x) = \x' (A - \alpha \hat I_S) \x \\
   \text{subject to} &  \mathbf{x} \in \Delta
\end{array}
\end{equation}
where $\hat I_S$ is the $n \times n$ diagonal matrix whose diagonal elements are 
set to 1 in correspondence to the vertices contained in $V \setminus S$ and to zero otherwise, and the 0's
represent null square matrices of appropriate dimensions.
In other words, assuming for simplicity that $S$ contains, say, the first $k$ vertices of $V$, we have:

$$
\hat I_S = 
\begin{pmatrix} 
  ~~0~~ & ~~0~~ \\ 
  ~~0~~ & I_{n-k}  
\end{pmatrix}
$$
where $I_{n-k}$ denotes the $(n-k) \times (n-k) $ principal submatrix of the $n \times n$ identity matrix $I$ 
indexed by the elements of $V \setminus S$.
Accordingly, the function $f_S^\alpha$ can also be written as follows:
$$
f_S^\alpha(\x) = \x' A \x - \alpha \x'_S \x_S
$$
$\x_S$ being the $(n-k)$-dimensional vector obtained from
$\x$ by dropping all the components in $S$.
Basically, the function $f_S^\alpha$ is obtained from $f$ by inserting in the
affinity matrix $A$ the value of the parameter $\alpha$ in the main diagonal positions
corresponding to the elements of $V \setminus S$.

Notice that this differs markedly, and indeed
generalizes, the formulation proposed in \cite{PavPel03} for obtaining a hierarchical clustering 
in that here, only a subset of elements in the main diagonal
is allowed to take the $\alpha$ parameter, the other ones being set to zero.
We note in fact that the original (non-regularized) dominant-set formulation (\ref{eq2}) \cite{PavPel07} 
as well as its regularized counterpart described in \cite{PavPel03}
can be considered as degenerate version of ours, corresponding to the cases $S=V$ and $S=\emptyset$, respectively.
It is precisely this increased flexibility which allows us to use
this idea for finding groups of ``constrained'' dominant-set clusters.

We now derive the Karush-Kuhn-Tucker (KKT) conditions for program (\ref{eqn:parQP}),
namely the first-order necessary conditions for local optimality (see, e.g., \cite{Lue84}).
For a point $\x \in \Delta$ to be a KKT-point there should exist $n$
nonnegative real constants $\mu_1 , \ldots , \mu_n$ and an additional real number $\lambda$
such that 
\begin{displaymath}
[(A - \alpha \hat I_S) \x]_i - \lambda + \mu_i = 0
\end{displaymath}
for all $i=1 \ldots n$, and
\begin{displaymath}
\sum_{i=1}^n x_i \mu_i = 0~.
\end{displaymath}
Since both the $x_i$'s and the $\mu_i$'s are nonnegative, the
latter condition is equivalent to saying that $i \in \sigma(\x)$ implies
$\mu_i= 0$, from which we obtain:
\begin{displaymath}
[(A - \alpha \hat I_S) \x]_i ~
\begin{cases} 
~ = ~   \lambda, ~ \mbox{ if } i \in \sigma(\x) \\ 
~ \le ~ \lambda, ~ \mbox{ if } i \notin \sigma(\x)  
\end{cases}
\end{displaymath}
for some constant $\lambda$.
Noting that $\lambda = \x'A\x - \alpha \x'_S \x_S$ and recalling the definition of $\hat I_S$,
the KKT conditions can be explicitly rewritten as:
\begin{equation}
\label{eqn:KKT}
\left\{
\begin{array}{llll}
(A\x)_i - \alpha x_i  & =   & \x'A\x - \alpha \x'_S \x_S, & \mbox{ if } i \in \sigma(\x) \mbox{ and } i \notin S \\
(A\x)_i               & =   & \x'A\x - \alpha \x'_S \x_S, & \mbox{ if } i \in \sigma(\x) \mbox{ and } i \in S \\
(A\x)_i               & \le & \x'A\x - \alpha \x'_S \x_S, & \mbox{ if } i \notin \sigma(\x)
\end{array}
\right.
\end{equation}

We are now in a position to discuss the main results which motivate the algorithm presented in this paper.
Note that, in the sequel, given a subset of vertices $S\subseteq V$, the face of $\Delta$ corresponding to $S$ is given by: $\Delta_{S}=\{x\in \Delta : \sigma (x)\subseteq S\}$.

\begin{proposition}
\label{prop:gamma}
Let $S \subseteq V$, with $S \neq \emptyset$.
Define
\begin{equation}
\label{eqn:defgamma}
\gamma_S = \max_{ \x \in \Delta_{V \setminus S}} 
\min_{i \in S} ~ \frac{\x' A \x - (A\x)_i}{\x' \x}
\end{equation}
and let $\alpha > \gamma_S$.
If $\x$ is a local maximizer of $f_S^\alpha$ in $\Delta$, then
$\sigma(\x) \cap S \neq \emptyset$.
\end{proposition}

\proof
Let $\x$ be a local maximizer of $f_S^\alpha$ in $\Delta$, and suppose
by contradiction that no element of $\sigma(\x)$ belongs to $S$ or, in other words,
that $\x \in \Delta_{V \setminus S}$.
By letting 
$$
i = \argmin_{j \in S} ~ \frac{\x' A \x - (A\x)_j}{\x' \x}
$$
and observing that $\sigma(\x) \subseteq V \setminus S$ implies $\x' \x = \x'_S \x_S$,
we have: 
$$
\alpha >  \gamma_S \ge 
\frac{\x' A \x - (A\x)_i}{\x' \x} =
\frac{\x' A \x - (A\x)_i}{\x'_S \x_S}~.
$$
Hence, $(A\x)_i > \x' A \x - \alpha \x'_S \x_S$ for $i \notin \sigma(\x)$, 
but this violates the KKT conditions (\ref{eqn:KKT}), thereby proving the proposition.
\endproof

The following proposition provides an easy-to-compute upper bound for $\gamma_S$. 

\begin{proposition}
\label{prop:bound}
Let $S \subseteq V$, with $S \neq \emptyset$. Then, 
\begin{equation}
\label{eqn:bound}
\gamma_S \le \lambda_{\max}(A_{V \setminus S})
\end{equation}
where $\lambda_{\max}(A_{V \setminus S})$ is the largest eigenvalue of the principal submatrix of $A$
indexed by the elements of $V \setminus S$. 
\end{proposition}
\proof
Let $\x$ be a point in $\Delta_{V \setminus S}$ which attains the maximum $\gamma_S$
as defined in (\ref{eqn:defgamma}).
Using the Rayleigh-Ritz theorem \cite{HorJon85} and the fact that $\sigma(\x)\subseteq V \setminus S$, we obtain:
$$
\lambda_{\max}(A_{V \setminus S}) \ge \frac{\x'_S A_{V \setminus S} \x_S}{\x'_S \x_S}
= \frac{\x' A \x}{\x' \x}~.
$$
Now, define $\gamma_S(\x) = \max \{ (A \x)_i~:~i \in S \}$.
Since $A$ is nonnegative so is $\gamma_S(\x)$, and recalling the definition of $\gamma_S$ we get:
$$
\frac{\x' A \x}{\x' \x} \ge \frac{\x' A \x - \gamma_S(\x)}{\x' \x} = \gamma_S
$$
which concludes the proof.
\endproof

The two previous propositions provide us with a simple technique to determine dominant-set clusters containing user-selected vertices. Indeed, if $S$ is the set of vertices selected by the user, by setting 
\begin{equation}
\label{alphabound}
\alpha > \lambda_{\max}(A_{V \setminus S})
\end{equation}
we are guaranteed that all local solutions of (\ref{eqn:parQP}) will have a support 
that necessarily contains elements of $S$.
As customary, we can use replicator dynamics or more sophisticated algorithms to find them.
Note that this does not necessarily imply that the (support of the) solution found corresponds to a dominant-set cluster of the original affinity matrix $A$, as adding the parameter $-\alpha$ on a portion of the main diagonal intrinsically changes the scale of the underlying problem. However, we have obtained extensive empirical evidence which supports a conjecture which turns out to be very useful for our interactive image segmentation application.

\begin{figure}[t]
\begin{center}
\includegraphics[scale=0.350]{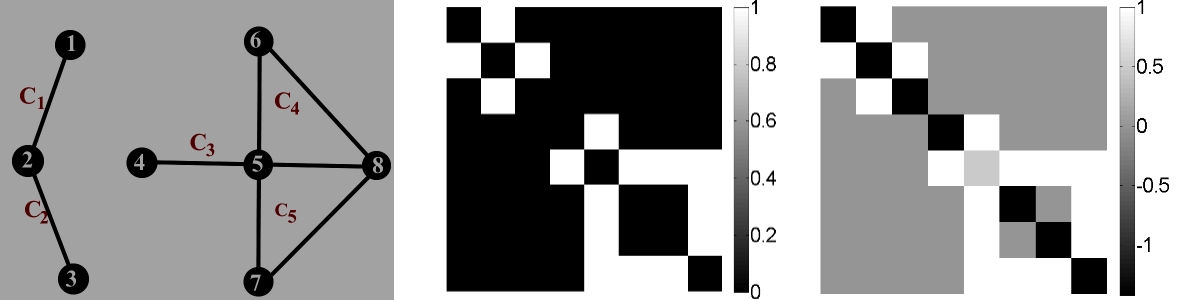}
\end{center}

\caption{\small An example graph (left), corresponding affinity matrix (middle), and scaled affinity matrix 
built considering vertex 5 as a user constraint (right). Notation $C_i$ refers to the $i^{th}$ maximal clique.}
\label{fig:ExamplarGraphAndAffinity}
\end{figure}

To illustrate the idea, let us consider the case where edge-weights are binary, which basically
means that the input graph is unweighted. In this case, it is known that dominant sets correspond to maximal cliques \cite{PavPel07}. Let $G=(V,E)$ be our unweighted graph and let $S$ be a subset of its vertices.
For the sake of simplicity, we distinguish three different situations of increasing generality.

\noindent
{\bf Case 1.} The set $S$ is a singleton, say $S = \{u\}$. In this case, we know from 
Proposition \ref{prop:bound} that all solutions $\x$ of 
$f_\alpha^S$ over $\Delta$ will have a support which contains $u$, that is $u \in \sigma(\x)$.
Indeed, we conjecture that there will be a unique local (and hence global) solution here whose support
coincides with the {\em union} of all maximal cliques of $G$ which contain vertex $u$. 

\noindent
{\bf Case 2.} The set $S$ is a clique, not necessarily maximal. In this case, 
Proposition \ref{prop:bound} predicts that all solutions $\x$ of (\ref{eqn:parQP})
will contain at least one vertex from $S$.
Here, we claim that indeed the support of local solutions is the union of the maximal cliques that contain $S$.

\noindent
{\bf Case 3.} The set $S$ is not a clique, but it can be decomposed as a collection of (possibly overlapping)
maximal cliques $C_1, C_2, ..., C_k$ (maximal with respect to the subgraph induced by $S$).
In this case, we claim that if $\x$ is a local solution, then its support can be obtained by taking the union of
all maximal cliques of $G$ containing one of the cliques $C_i$ in $S$.

To make our discussion clearer, consider the graph shown in Fig. \ref{fig:ExamplarGraphAndAffinity}. 
In order to test whether our claims hold, we used as the set $S$ different combinations of vertices, and
enumerated all local solutions of (\ref{eqn:parQP}) by multi-start replicator dynamics.
Some results are shown below, where on the left-hand side we indicate the set $S$, while on
the right hand-side we show the supports provided as output by the different runs of the algorithm. 
\begin{table}[h]
\begin{tabular}
{p{2.2cm}  p{0.5cm}  p{6cm}}
1.  ~~$S = \{ 2 \}$       &   $\Rightarrow$     & $\sigma(\x) = \{ 1, 2, 3 \}$ \\
2.  ~~$S = \{ 5 \}$       &   $\Rightarrow$     & $\sigma(\x) = \{ 4, 5, 6, 7, 8 \}$ \\
3.  ~~$S = \{ 4, 5 \}$    &   $\Rightarrow$     & $\sigma(\x) = \{ 4, 5 \}$ \\
4.  ~~$S = \{ 5, 8 \}$    &   $\Rightarrow$     & $\sigma(\x) = \{ 5, 6, 7, 8 \}$ \\
5.  ~~$S = \{ 1, 4 \}$    &   $\Rightarrow$     & $\sigma(\x_1) = \{ 1, 2 \}$, ~~$\sigma(\x_2) = \{ 4, 5 \}$\\
6.  ~~$S = \{ 2, 5, 8 \}$ &   $\Rightarrow$     & $\sigma(\x_1) = \{ 1, 2, 3 \}$, ~~$\sigma(\x_2) = \{ 5, 6, 7, 8 \}$
\end{tabular}
\end{table}

The previous observations can be summarized in the following general statement which does comprise all three cases. 
Let $S = C_1 \cup C_2 \cup \ldots \cup C_k$ ($k \ge 1$) be a subset of vertices of $G$, 
consisting of a collection of cliques $C_i$ ($i=1 \ldots k$).
Suppose that condition (\ref{alphabound}) holds, and let 
$\x$ be a local solution of (\ref{eqn:parQP}). Then, $\sigma(\x)$ consists of the union of 
all maximal cliques containing some clique $C_i$ of $S$.

We conjecture that the previous claim carries over to edge-weighted graphs,
where the notion of a maximal clique is replaced by that of a dominant set.
In the supplementary material we report the results of an extensive experimentation
we have conducted on standard DIMACS graphs which provide support to our claim.
This is going to play a key role in our applications of these ideas
to interactive image segmentation.

\section{Application to interactive image segmentation}

In this section we apply our model to the interactive image segmentation problem. As input modalities we consider scribbles as well as boundary-based approaches (in particular, bounding boxes) and, in both cases,
we show how the system is robust under input perturbations, namely imperfect scribbles or loose bounding boxes.

In this application the vertices of the underlying graph $G$ represent the pixels of the input image (or superpixels, as discussed below), and the edge-weights reflect the similarity between them.
As for the set $S$, its content depends on whether we are using scribbles or bounding boxes 
as the user annotation modality. 
In particular, in the case of scribbles, $S$ represents precisely those pixels that have been
manually selected by the user. In the case of boundary-based annotation 
instead, it is taken to contain only the pixels comprising the box boundary,
which are supposed to represent the background scene.
Accordingly, the union of the extracted dominant sets, say $\mathcal{L}$ dominant sets are extracted which contain the set $S$,  as described in the previous section and below, $\mathbf{UDS}=\mathcal{D}_1 \cup \mathcal{D}_2 ..... \cup \mathcal{D}_{\mathcal{L}}$, represents either the foreground object or the background scene depending on the input modality. For scribble-based approach the extracted set, $\mathbf{UDS}$, represent the segmentation result, while in the boundary-based approach we provide as output the complement of the extracted set, namely $\mathbf{V}\setminus \mathbf{UDS}$. 

\begin{figure}[t]
\centering
\includegraphics[scale=0.1]{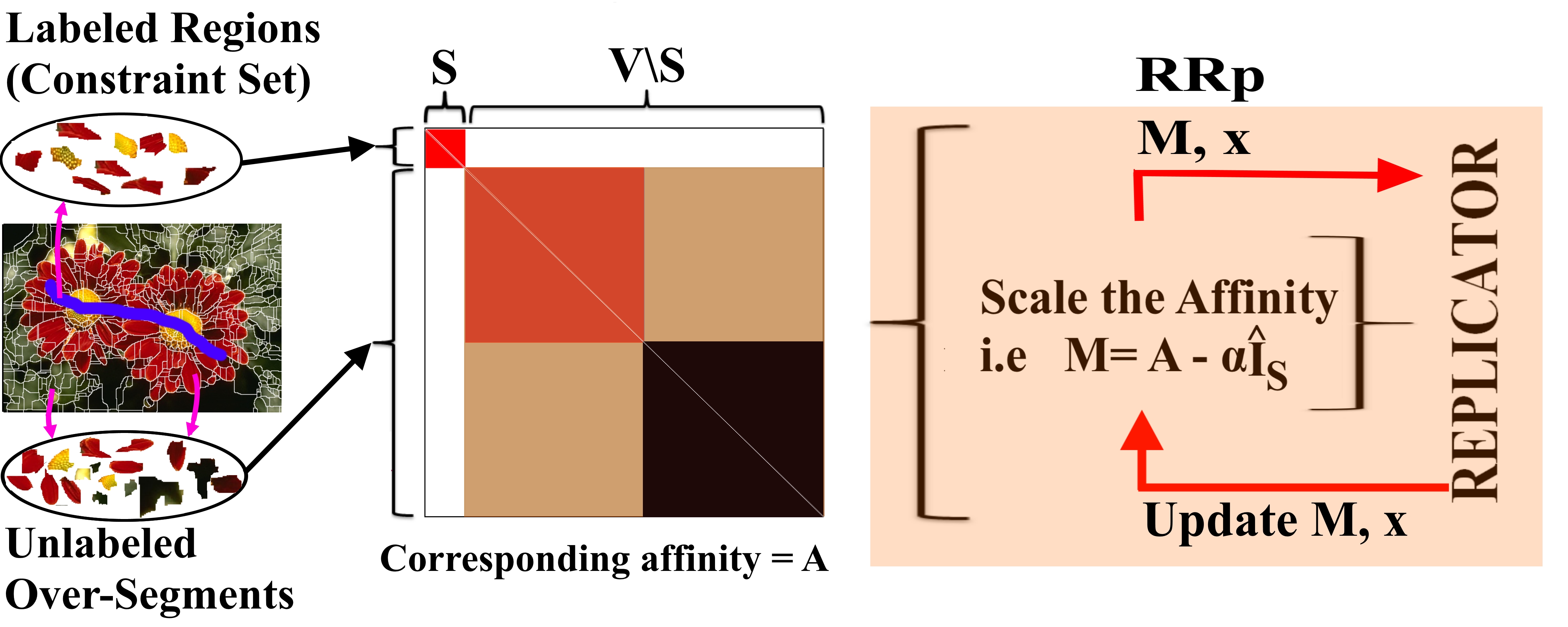}
\caption{\small Overview of our system. \textbf{Left:} Over-segmented image (output of the UCM-OWT algorithm \cite{MalikHierarchical}) with a user scribble (blue label). \textbf{Middle:} The corresponding affinity matrix, using each over-segments as a node, showing its two parts: $S$, the constraint set which contains the user labels, and $V\setminus S$, the part of the graph which takes the regularization parameter $\alpha$. \textbf{Right:} RRp, starts from the barycenter and extracts the first dominant set and update $\mathbf{x}$ and $\mathbf{M}$, for the next extraction till all the dominant sets which contain the user labeled regions are extracted.}
\label{fig:Framework}
\end{figure}

Figure \ref{fig:Framework} shows the pipeline of our system. 
Many segmentation tasks reduce their complexity by using superpixels (a.k.a. over-segments) as a preprocessing step \cite{MilCutCVPR14,LOOSECUTcorr15,HoiEfrHebICCV2005,WanJiaHuaZhaQuaCVPR2008,XiaQuaICCV2009}. While \cite{MilCutCVPR14} used SLIC superpixels \cite{SLIC-superpixels-TPAMI12,LOOSECUTcorr15} used a recent superpixel algorithm \cite{ZhoJuWanWACV2015} which considers not only the color/feature information but also boundary smoothness among the superpixels. 
In this work, we used the over-segments obtained from Ultrametric Contour Map (UCM) which is constructed from Oriented Watershed Transform (OWT) using globalized probability of boundary (gPb) signal as an input \cite{MalikHierarchical}. 

We then construct a graph $G$ where the vertices represent over-segments and 
the similarity (edge-weight) between any two of them is obtained using a standard Gaussian kernel
$$\mat A^\sigma_{ij}=\ind{i\neq j}exp(\Vert\vct f_i-\vct f_j\Vert^2/{2\sigma^2})$$
where $\vct f_i$, is the feature vector of the $i^{th}$ over-segment, $\sigma$ is the free scale parameter, and $\ind{P}=1$ if $P$ is true, $0$ otherwise.

Given the affinity matrix $A$ and the set $S$ as described before, the system constructs
the regularized matrix $M=A-\alpha \hat I_S$, with $\alpha$ chosen as prescribed in (\ref{alphabound}).
Then, the replicator dynamics (\ref{eqn:Replicator}) are run (starting them as customary from the simplex barycenter) until they converge to some solution vector $\x$. We then take the support of $\x$, remove the
corresponding vertices from the graph and restart the replicator dynamics until all
the elements of $S$ are extracted.

\subsection{Experiments and results}

As mentioned above, the vertices of our graph represents over-segments
and edge weights (similarities) are built from the median of the color of all pixels in RGB, HSV, and L*a*b* color spaces, and Leung-Malik (LM) Filter Bank \cite{MalikLeungLMfilterIJCV01}. The number of dimensions of feature vectors for each over-segment is then 57 (three for each of the RGB, L*a*b*, and HSV color spaces, and 48 for LM Filter Bank).

In practice, the performance of graph-based algorithms that use Gaussian kernel, as we do, is sensitive to the selection of the scale parameter $\sigma$. In our experiments, we have reported three different results based on the way $\sigma$ is chosen: $1)$ CDS\_Best\_Sigma, in this case the best parameter $\sigma$  is selected on a per-image basis, which indeed can be thought of as the optimal result (or upper bound) of the framework. $2) $ CDS\_Single\_Sigma, the best parameter in this case is selected on a per-database basis tuning $\sigma$ in some fixed range, which in our case is between 0.05 and 0.2. $3)$ CDS\_Self\_Tuning, the $\sigma^2$ in the above equation is replaced, based on \cite{ManPieNIPS2004}, by $\sigma_i*\sigma_j$, where $\sigma_i = mean(KNN(f_i))$, the mean of the K\_Nearest\_Neighbor of the sample $f_i$, K is fixed in all the experiment as 7. 

\textbf{Datasets:} We conduct four different experiments on the well-known GrabCut dataset \cite{GrabCutRotherKB04} which has been used as a benchmark in many computer vision tasks \cite{LiMN13,iccvLempitsky09,TangECCV14,OneCutICCV13,MilCutCVPR14,LOOSECUTcorr15,PriMorCohCVPR2010,YanCaiZheLuoIP2010}. The dataset contains 50 images together with manually-labeled segmentation ground truth. The same bounding boxes as those in \cite{iccvLempitsky09} is used as a baseline bounding box. We also evaluated our scribbled-based approach using the well known Berkeley dataset which contains 100 images.

\textbf{Metrics:}
We evaluate the approach using different metrics: error rate, fraction of misclassified pixels within the bounding box, Jaccard index which is given by, following \cite{EvaluationMetricsMcGuinnessO10}, $J$ = $\frac{|GT \cap O|}{|GT \cup O|}$, where $GT$ is the ground truth and $O$ is the output. The third metric is the Dice Similarity Coefficient ($DSC$), which measures the overlap between two segmented object volume, and is computed as $DSC = \frac{2*|GT \cap O|}{|GT|+|O|}$.

\textbf{Annotations:} 
In interactive image segmentation, users provide annotations which guides the segmentation.
A user usually provides information in different forms such as scribbles and bounding boxes. The input modality affects both its accuracy and ease-of-use \cite{JainGraICCV2013}. However, existing methods fix themselves to one input modality and focus on how to use that input information effectively. This leads to a suboptimal tradeoff in user and machine effort. Jain et al. \cite{JainGraICCV2013} estimates how much user input is required to sufficiently segment a given image. In this work, as we have proposed an interactive framework, figure \ref{fig:InputModalities}, which can take any type of input modalities, we will use four different types of annotations: bounding box, loose bounding box, scribbles - only on the object of interest -, and scribbles with error as of \cite{BaiWuCVPR2014}.

\subsubsection{Scribble based segmentation} \label{Sec.Scribble}
Given labels on the foreground as constraint set, we built the graph and collect (iteratively) all unlabeled regions (nodes of the graph) by extracting dominant set(s) that contains the constraint set (user scribbles). We provided quantitative comparison against several recent state-of-the-art interactive image segmentation methods which uses scribbles as a form of human annotation: \cite{BoyJolICCV2001}, Lazy Snapping \cite{LiSunTanShuACM2004}, Geodesic Segmentation \cite{BaiSapIJCV2009}, Random Walker \cite{GraPAMI2006}, Transduction \cite{DucAudKerPonFloCVPR2008}, Geodesic Graph Cut \cite{PriMorCohCVPR2010}, Constrained Random Walker \cite{YanCaiZheLuoIP2010}.
Tables \ref{table:ScribblesResult},\ref{table:ScribblesResultBerkeley} and the plots in Figure \ref{fig:ExamplarResults} show the respective quantitative and the several qualitative segmentation results. Most of the results, reported on table \ref{table:ScribblesResult}, are reported by previous works \cite{LOOSECUTcorr15,MilCutCVPR14,iccvLempitsky09,PriMorCohCVPR2010,YanCaiZheLuoIP2010}.
We can see that the proposed framework outperforms all the other approaches.

\begin{table}[t]
\parbox{.45\linewidth}{
\centering
\begin{tabular}{l|r}

Methods                                        & Error Rate \\ \hline
Graph Cut \cite{BoyJolICCV2001}                & 6.7 \\ \hline
Lazy Snapping \cite{LiSunTanShuACM2004}        & 6.7 \\ \hline
Geodesic Segmentation \cite{BaiSapIJCV2009}    & 6.8 \\ \hline
Random Walker \cite{GraPAMI2006}               & 5.4 \\ \hline
Transduction \cite{DucAudKerPonFloCVPR2008}    & 5.4 \\ \hline
Geodesic Graph Cut \cite{PriMorCohCVPR2010}    & 4.8 \\ \hline
Constrained Random Walker \cite{YanCaiZheLuoIP2010} & 4.1 \\ \hline
CDS\_Self Tuning (Ours)               & \textbf{3.57} \\ \hline
CDS\_Single Sigma (Ours)              & \textbf{3.80} \\ \hline
CDS\_Best Sigma (Ours)                & 2.72 \\ \hline
\end{tabular}
\caption{\small Error rates of different scribble-based approaches on the Grab-Cut dataset.}
\label{table:ScribblesResult}
}
\hfill
\parbox{.45\linewidth}{
\centering
\begin{tabular}{l|r}
Methods                          & Jaccard Index \\ \hline
MILCut-Struct \cite{MilCutCVPR14}             & 84 \\ \hline
MILCut-Graph \cite{MilCutCVPR14}              & 83 \\ \hline
MILCut \cite{MilCutCVPR14}                    & 78 \\ \hline
Graph Cut \cite{GrabCutRotherKB04}            & 77 \\ \hline
Binary Partition Trees \cite{SalGarIP2000}    & 71 \\ \hline
Interactive Graph Cut \cite{BoyJolICCV2001}   & 64 \\ \hline
Seeded Region Growing \cite{AdaBisPAMI1994}   & 59 \\ \hline
Simple Interactive O.E\cite{FriJanRojACM2005} & 63 \\ \hline
CDS\_Self Tuning (Ours)                       & \textbf{93} \\ \hline
CDS\_Single Sigma (Ours)                      & \textbf{93} \\ \hline
CDS\_Best Sigma (Ours)                        & 95 \\ \hline
\end{tabular}
\caption{\small Jaccard Index of different approaches -- first 5 bounding-box-based -- on Berkeley dataset.}
\label{table:ScribblesResultBerkeley}
}
\end{table}
\textbf{Error-tolerant Scribble Based Segmentation:} This is a family of scribble-based approach, proposed by Bai et. al \cite{BaiWuCVPR2014}, which tolerates imperfect input scribbles thereby avoiding the assumption of accurate scribbles. We have done experiments using synthetic scribbles and compared the algorithm against recently proposed methods specifically designed to segment and extract the object of interest tolerating the user input errors \cite{BaiWuCVPR2014,LiuSunShuACM2009,OzaKemAydACM2012,SubParSolKauCGF2013}.

Our framework is adapted to this problem as follows. We give, for the framework, the foreground scribbles as constraint set and check those scribbled regions which include background scribbled regions as their members in the extracted dominant set. Collecting all those dominant sets which are free from background scribbled regions generates the object of interest.

\begin{figure}[t]
\centering
\includegraphics[scale=0.12]{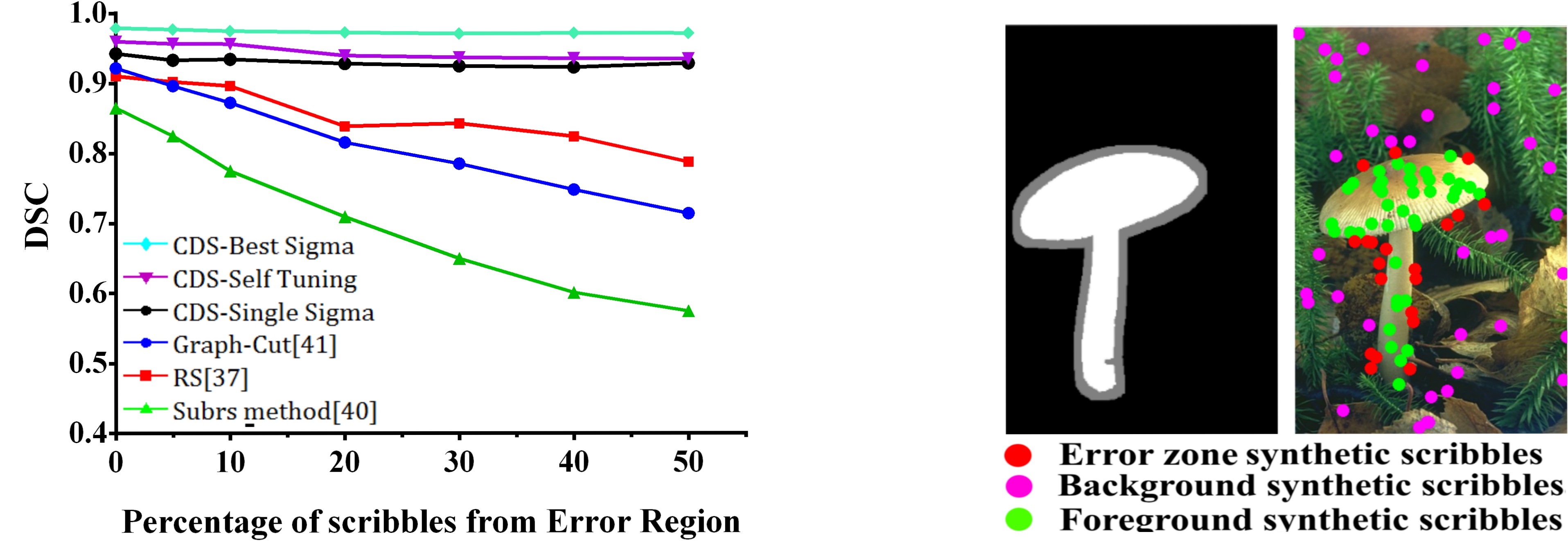}
\caption{\small  \textbf{Left:} Performance of algorithms, on Grab-Cut dataset, for different percentage of synthetic scribbles from the error region. \textbf{Right:} Synthetic scribbles and error region }
\label{fig:SyntheticScribblesResult}
\end{figure}

\textbf{Experiment using synthetic scribbles.}
Here, a procedure similar to the one used in \cite{SubParSolKauCGF2013} and \cite{BaiWuCVPR2014} has been followed.
First, 50 foreground pixels and 50 background pixels are randomly selected based on ground truth (see Fig. \ref{fig:SyntheticScribblesResult}). They are then assigned as foreground or background scribbles, respectively. Then
an error-zone for each image is defined as background pixels that are less than a distance D from the foreground, in
which D is defined as 5 \%. We randomly select 0 to 50 pixels in the error zone and assign them as foreground scribbles to simulate different degrees of user input errors. We randomly select 0, 5, 10, 20, 30, 40, 50
erroneous sample pixels from error zone to simulate the error percentage of 0\%, 10\%, 20\%, 40\%, 60\%, 80\%, 100\%
in the user input. It can be observed from figure \ref{fig:SyntheticScribblesResult} that our approach is not affected by the increase in the percentage of scribbles from error region.

\subsubsection{Segmentation using bounding boxes} \label{Sec.boundingBox}
The goal here is to segment the object of interest out from the background based on a given bounding box. The corresponding over-segments which contain the box label are taken as constraint set which guides the segmentation. The union of the extracted set is then considered as background while the union of other over-segments represent the object of interest.

\begin{figure}[t]
\centering
\includegraphics[height=16cm,width=12cm]{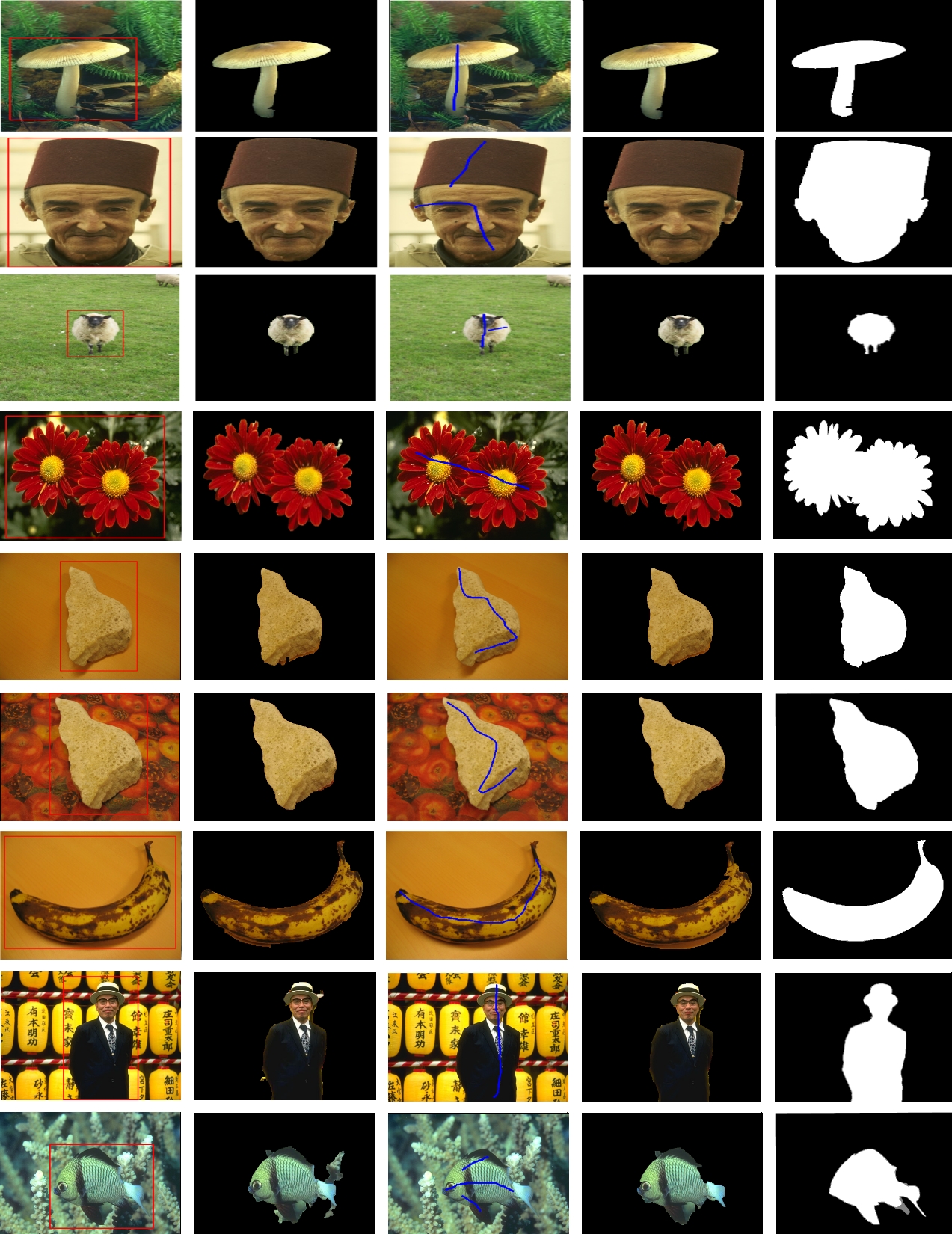}
\caption{\small Examplar results of the algorithm tested on Grab-Cut dataset. \textbf{Left:} Original image with bounding boxes of \cite{iccvLempitsky09}. \textbf{Middle left:} Result of the bounding box approach. \textbf{Middle:} Original image and scribbles (observe that the scribles are only on the object of interest). \textbf{Middle right:} Results of the scribbled approach. \textbf{Right:} The ground truth.}
\label{fig:ExamplarResults}
\end{figure}

We provide quantitative comparison against several recent state-of-the-art interactive image segmentation methods which uses bounding box: LooseCut \cite{LOOSECUTcorr15}, GrabCut \cite{GrabCutRotherKB04}, OneCut \cite{OneCutICCV13}, MILCut \cite{MilCutCVPR14}, pPBC and \cite{TangECCV14}.
Table \ref{table:LoosCutApproach} and the pictures in Figure \ref{fig:ExamplarResults} show the respective error rates and the several qualitative segmentation results. Most of the results, reported on table \ref{table:LoosCutApproach}, are reported by previous works \cite{LOOSECUTcorr15,MilCutCVPR14,iccvLempitsky09,PriMorCohCVPR2010,YanCaiZheLuoIP2010}.

\textbf{Segmentation Using Loose Bounding Box:} This is a variant of the bounding box approach, proposed by Yu et.al \cite{LOOSECUTcorr15}, which avoids the dependency of algorithms on the tightness of the box enclosing the object of interest. The approach not only avoids the annotation burden but also allows the algorithm to use automatically detected bounding boxes which might not tightly encloses the foreground object. It has been shown, in \cite{LOOSECUTcorr15}, that the well-known GrabCut algorithm \cite{GrabCutRotherKB04} fails when the looseness of the box is increased. Our framework, like \cite{LOOSECUTcorr15}, is able to extract the object of interest in both tight and loose boxes. Our algorithm is tested against a series of bounding boxes with increased looseness. The bounding boxes of \cite{iccvLempitsky09} are used as boxes with 0\% looseness. A looseness $L$ (in percentage) means an increase in the area of the box against the baseline one. The looseness is increased, unless it reaches the image perimeter where the box is cropped, by dilating the box by a number of pixels, based on the percentage of the looseness, along the 4 directions: left, right, up, and down.

For the sake of comparison, we conduct the same experiments as in \cite{LOOSECUTcorr15}: 41 images out of the 50 GrabCut dataset \cite{GrabCutRotherKB04} are selected as the rest 9 images contain multiple objects while the ground truth is only annotated on a single object. As other objects, which are not marked as an object of interest in the ground truth,  may be covered when the looseness of the box increases, images of multiple objects are not applicable
for testing the loosely bounded boxes \cite{LOOSECUTcorr15}. Table \ref{table:LoosCutApproach} summarizes the results of different approaches using bounding box at different level of looseness. As can be observed from the table, our approach performs well compared to the others when the level of looseness gets increased. When the looseness $L=0$, \cite{MilCutCVPR14} outperforms all, but it is clear, from their definition of tight bounding box, that it is highly dependent on the tightness of the bounding box. It even shrinks the initially given bounding box by 5\% to ensure its tightness before the slices of the positive bag are collected. For looseness of $L=120$ we have similar result with LooseCut \cite{LOOSECUTcorr15} which is specifically designed for this purpose. For other values of $L$ our algorithm outperforms all the approaches.
\begin{table}[t]
  \centering
\begin{tabular}
{p{3.4cm}    | p{1.81cm} | p{1.81cm}| p{1.81cm}| p{1.81cm}}
\hline\hline
Methods      & $ L = 0\% $ & $ L = 120\% $ & $ L = 240\% $ & $ L = 600\% $  \\ \hline
GrabCut \cite{GrabCutRotherKB04}      & 7.4 &  10.1 &  12.6     &  13.7   \\ \hline
OneCut \cite{OneCutICCV13}            & 6.6 & 8.7  &  9.9      &  13.7  \\ \hline
pPBC \cite{TangECCV14}                & 7.5  &  9.1  &  9.4      &  12.3  \\ \hline
MilCut \cite{MilCutCVPR14}            & \textbf{3.6}&  -    &  -        &  -       \\ \hline
LooseCut \cite{LOOSECUTcorr15}        & 7.9 & \textbf{5.8}  &  6.9      &  6.8     \\ \hline
CDS\_Self Tuning (Ours)               & 7.54 &  6.78 &  \textbf{6.35}      &  7.17 \\ \hline
CDS\_Single Sigma (Ours)              & 7.48 &  5.9  &  \textbf{6.32}      & \textbf{6.29} \\ \hline
CDS\_Best Sigma (Ours)                & 6.0  &  4.4  &  4.2     &  4.9 \\ \hline
\end{tabular}
\caption{\small Error rates of different boundin-box approaches with different level of looseness as an input, on the Grab-Cut dataset. $ L = 0\% $ implies a baseline bounding box as those in \cite{iccvLempitsky09}}
\label{table:LoosCutApproach}
\end{table}

\textbf{Complexity:} 
In practice, over-segmenting and extracting features may be treated as a pre-processing step which can be done
before the segmentation process. Given the affinity matrix, a simple and effective optimization algorithm to extract the object of interest is given by the replicator dynamics \ref{eqn:Replicator}. Its computational complexity per step is $O(N^2)$, with N being the total number of nodes of the graph. Infection-immunization dynamics \cite{RotBomGEB2011} is a faster alternative which has an $O(N)$ complexity for each step which allow convergence of the framework in fraction of second, with a code written in Matlab and run on a core i5 6 GB of memory. As for the pre-processing step, the original \textit{gPb-owt-ucm} segmentation algorithm was very slow to be used as a practical tools. Catanzaro et al. \cite{CatSuSunLeeMurKeuICCV2009} proposed a faster alternative, which reduce the runtime from 4 minutes to 1.8 seconds, reducing the  computational complexity and using parallelization which allow \textit{gPb} contour detector and \textit{gPb-owt-ucm} segmentation algorithm practical tools. For the purpose of our experiment we have used the Matlab implementation which takes around four minutes to converge, but in practice it is possible to give for our framework as an input, the GPU implementation \cite{CatSuSunLeeMurKeuICCV2009} which allows the convergence of the whole framework in around 4 seconds.

\section{Conclusions}

In this paper, we have developed an interactive image segmentation algorithm based on 
the idea of finding a collection of dominant-set clusters constrained to contain 
the elements of a user annotation.
The approach is based on some properties of a family of quadratic optimization problems
related to dominant sets which show that, by properly selecting a regularization parameter
that controls the structure of the underlying function,
we are able to ``force'' all solutions to contain the user-provided elements.
The resulting algorithm is capable of dealing with both scribble-based and boundary-based annotation modes.
Segmentation results of extensive experiments on natural images demonstrate that the approach 
compares favorably with state-of-the-art algorithms and turns out to be robust 
in the presence of loose bounding boxes and large amount of user input errors.
Future work will focus on applying the framework on video sequences and other computer vision problems
such as content-based image retrieval.

\medskip
\noindent
{\bf Acknowledgments.} This work has been partly supported by Samsung Global Research Outreach Program.

\clearpage
\bibliographystyle{splncs}
\bibliography{1636}

\begin{thebibliography}{10}

\bibitem{GrabCutRotherKB04}
Rother, C., Kolmogorov, V., Blake, A.:
\newblock ``{G}rabcut'': {I}nteractive foreground extraction using iterated
  graph cuts.
\newblock {ACM} Trans. Graph. \textbf{23}(3) (2004)  309--314

\bibitem{iccvLempitsky09}
Lempitsky, V.S., Kohli, P., Rother, C., Sharp, T.:
\newblock Image segmentation with a bounding box prior.
\newblock In: {ICCV}. (2009)  277--284

\bibitem{MilCutCVPR14}
Wu, J., Zhao, Y., Zhu, J., Luo, S., Tu, Z.:
\newblock Milcut: {A} sweeping line multiple instance learning paradigm for
  interactive image segmentation.
\newblock In: {CVPR}. (2014)  256--263

\bibitem{BaiSapIJCV2009}
Bai, X., Sapiro, G.:
\newblock Geodesic matting: {A} framework for fast interactive image and video
  segmentation and matting.
\newblock Int. J. Computer Vision \textbf{82}(2) (2009)  113--132

\bibitem{LiSunTanShuACM2004}
Li, Y., Sun, J., Tang, C., Shum, H.:
\newblock Lazy snapping.
\newblock {ACM} Trans. Graph. \textbf{23}(3) (2004)

\bibitem{ProSapIP2007}
Protiere, A., Sapiro, G.:
\newblock Interactive image segmentation via adaptive weighted distances.
\newblock {IEEE} Trans. Image Processing \textbf{16}(4) (2007)  1046--1057

\bibitem{BoyJolICCV2001}
Boykov, Y., Jolly, M.:
\newblock Interactive graph cuts for optimal boundary and region segmentation
  of objects in {N-D} images.
\newblock In: {ICCV}. (2001)  105--112

\bibitem{MorBarIP1998}
Mortensen, E.N., Barrett, W.A.:
\newblock Interactive segmentation with intelligent scissors.
\newblock Graphical Models and Image Processing \textbf{60}(5) (1998)  349--384

\bibitem{LOOSECUTcorr15}
Yu, H., Zhou, Y., Qian, H., Xian, M., Lin, Y., Guo, D., Zheng, K., Abdelfatah,
  K., Wang, S.:
\newblock Loosecut: Interactive image segmentation with loosely bounded boxes.
\newblock CoRR \textbf{abs/1507.03060} (2015)

\bibitem{XiaZhaCheXuDinCoRR2015}
Xian, M., Zhang, Y., Cheng, H.D., Xu, F., Ding, J.:
\newblock Neutro-connectedness cut.
\newblock CoRR \textbf{abs/1512.06285}

\bibitem{BaiWuCVPR2014}
Bai, J., Wu, X.:
\newblock Error-tolerant scribbles based interactive image segmentation.
\newblock In: {CVPR}. (2014)  392--399

\bibitem{JainGraICCV2013}
Jain, S.D., Grauman, K.:
\newblock Predicting sufficient annotation strength for interactive foreground
  segmentation.
\newblock In: {ICCV}. (2013)  1313--1320

\bibitem{PavPelCVPR2003}
Pavan, M., Pelillo, M.:
\newblock A new graph-theoretic approach to clustering and segmentation.
\newblock In: CVPR. (2003)  145--152

\bibitem{PavPel07}
Pavan, M., Pelillo, M.:
\newblock Dominant sets and pairwise clustering.
\newblock IEEE Trans. Pattern Anal. Mach. Intell. \textbf{29}(1) (2007)
  167--172

\bibitem{RotPelPAMI2013}
{Rota Bul{\`{o}}}, S., Pelillo, M.:
\newblock A game-theoretic approach to hypergraph clustering.
\newblock {IEEE} Trans. Pattern Anal. Mach. Intell. \textbf{35}(6) (2013)
  1312--1327

\bibitem{PavPel03}
Pavan, M., Pelillo, M.:
\newblock Dominant sets and hierarchical clustering.
\newblock In: ICCV. (2003)  362--369

\bibitem{Lue84}
Luenberger, D.G., Ye, Y.:
\newblock Linear and Nonlinear Programming.
\newblock Springer, New York (2008)

\bibitem{HorJon85}
Horn, R.A., Johnson, C.R.:
\newblock Matrix Analysis.
\newblock Cambridge University Press, New York (1985)

\bibitem{HoiEfrHebICCV2005}
Hoiem, D., Efros, A.A., Hebert, M.:
\newblock Geometric context from a single image.
\newblock In: ICCV. (2005)  654--661

\bibitem{WanJiaHuaZhaQuaCVPR2008}
Wang, J., Jia, Y., Hua, X., Zhang, C., Quan, L.:
\newblock Normalized tree partitioning for image segmentation.
\newblock In: CVPR. (2008)

\bibitem{XiaQuaICCV2009}
Xiao, J., Quan, L.:
\newblock Multiple view semantic segmentation for street view images.
\newblock In: ICCV. (2009)  686--693

\bibitem{SLIC-superpixels-TPAMI12}
Achanta, R., Shaji, A., Smith, K., Lucchi, A., Fua, P., S{\"{u}}sstrunk, S.:
\newblock {SLIC} superpixels compared to state-of-the-art superpixel methods.
\newblock {IEEE} Trans. Pattern Anal. Mach. Intell. \textbf{34}(11) (2012)
  2274--2282

\bibitem{ZhoJuWanWACV2015}
Zhou, Y., Ju, L., Wang, S.:
\newblock Multiscale superpixels and supervoxels based on hierarchical
  edge-weighted centroidal voronoi tessellation.
\newblock In: WACV. (2015)  1076--1083

\bibitem{MalikHierarchical}
Arbelaez, P., Maire, M., Fowlkes, C.C., Malik, J.:
\newblock Contour detection and hierarchical image segmentation.
\newblock {IEEE} Trans. Pattern Anal. Mach. Intell. \textbf{33} (2011)
  898--916

\bibitem{MalikLeungLMfilterIJCV01}
Leung, T.K., Malik, J.:
\newblock Representing and recognizing the visual appearance of materials using
  three-dimensional textons.
\newblock Int. J. Computer Vision \textbf{43}(1) (2001)  29--44

\bibitem{ManPieNIPS2004}
Zelnik-Manor, L., Perona, P.:
\newblock Self-tuning spectral clustering.
\newblock In: NIPS. (2004)  1601--1608

\bibitem{LiMN13}
Li, H., Meng, F., Ngan, K.N.:
\newblock Co-salient object detection from multiple images.
\newblock {IEEE} Trans. Multimedia \textbf{15}(8) (2013)  1896--1909

\bibitem{TangECCV14}
Tang, M., Ayed, I.B., Boykov, Y.:
\newblock Pseudo-bound optimization for binary energies.
\newblock In: {ECCV}. (2014)  691--707

\bibitem{OneCutICCV13}
Tang, M., Gorelick, L., Veksler, O., Boykov, Y.:
\newblock Grabcut in one cut.
\newblock In: ICCV. (2013)  1769--1776

\bibitem{PriMorCohCVPR2010}
Price, B.L., Morse, B.S., Cohen, S.:
\newblock Geodesic graph cut for interactive image segmentation.
\newblock In: CVPR. (2010)  3161--3168

\bibitem{YanCaiZheLuoIP2010}
Yang, W., Cai, J., Zheng, J., Luo, J.:
\newblock User-friendly interactive image segmentation through unified
  combinatorial user inputs.
\newblock IEEE Trans. Image Processing \textbf{19}(9) (2010)  2470--2479

\bibitem{EvaluationMetricsMcGuinnessO10}
McGuinness, K., O'Connor, N.E.:
\newblock A comparative evaluation of interactive segmentation algorithms.
\newblock Pattern Recognition \textbf{43}(2) (2010)  434--444

\bibitem{GraPAMI2006}
Grady, L.:
\newblock Random walks for image segmentation.
\newblock {IEEE} Trans. Pattern Anal. Mach. Intell. \textbf{28}(11) (2006)
  1768--1783

\bibitem{DucAudKerPonFloCVPR2008}
Duchenne, O., Audibert, J., Keriven, R., Ponce, J., S{\'{e}}gonne, F.:
\newblock Segmentation by transduction.
\newblock In: CVPR. (2008)

\bibitem{SalGarIP2000}
Salembier, P., Garrido, L.:
\newblock Binary partition tree as an efficient representation for image
  processing, segmentation, and information retrieval.
\newblock {IEEE} Trans. Image Processing \textbf{9}(4) (2000)  561--576

\bibitem{AdaBisPAMI1994}
Adams, R., Bischof, L.:
\newblock Seeded region growing.
\newblock {IEEE} Trans. Pattern Anal. Mach. Intell. \textbf{16}(6) (1994)
  641--647

\bibitem{FriJanRojACM2005}
Friedland, G., Jantz, K., Rojas, R.:
\newblock {SIOX:} {S}imple interactive object extraction in still images.
\newblock In: {(ISM}. (2005)  253--260

\bibitem{LiuSunShuACM2009}
Liu, J., Sun, J., Shum, H.:
\newblock Paint selection.
\newblock {ACM} Trans. Graph. \textbf{28}(3) (2009)

\bibitem{OzaKemAydACM2012}
Sener, O., Ugur, K., Alatan, A.A.:
\newblock Error-tolerant interactive image segmentation using dynamic and
  iterated graph-cuts.
\newblock In: IMMPD@ACM Multimedia. (2012)  9--16

\bibitem{SubParSolKauCGF2013}
Subr, K., Paris, S., Soler, C., Kautz, J.:
\newblock Accurate binary image selection from inaccurate user input.
\newblock Comput. Graph. Forum \textbf{32}(2) (2013)  41--50

\bibitem{RotBomGEB2011}
{Rota Bul{\`{o}}}, S., Bomze, I.M.:
\newblock Infection and immunization: {A} new class of evolutionary game
  dynamics.
\newblock Games and Economic Behavior \textbf{71}(1) (2011)  193--211

\bibitem{CatSuSunLeeMurKeuICCV2009}
Catanzaro, B.C., Su, B., Sundaram, N., Lee, Y., Murphy, M., Keutzer, K.:
\newblock Efficient, high-quality image contour detection.
\newblock In: {ICCV}. (2009)  2381--2388

\end{thebibliography}
\end{document}